\documentclass[letterpaper, 10 pt, conference]{ieeeconf}  
\IEEEoverridecommandlockouts                              

\usepackage{graphicx}
\usepackage{amsmath}
\usepackage{amssymb}
\usepackage{booktabs}
\usepackage{multirow}
\usepackage{tabularx}
\usepackage{xcolor}
\usepackage{tikz}
\usepackage{bm}
\usepackage{arydshln}
\usetikzlibrary{arrows.meta,positioning,fit}
\usepackage{overpic}

\definecolor{methodblue}{RGB}{46,109,164}
\definecolor{methodgreen}{RGB}{53,142,83}
\definecolor{methodorange}{RGB}{221,132,40}
\definecolor{methodred}{RGB}{187,64,64}
\definecolor{placeholder}{RGB}{242,242,242}
\newcommand{\method}{\textsc{FaRe}}

\newcommand{\ie}{\textit{i}.\textit{e}.}

\newcommand{\cf}{\textit{cf}.}

\title{\LARGE \bf
Causal-History Test-Time Scaling for Failure Recovery in Autoregressive World-Action Models
}

\author{
Lin Li,
Long Chen,
Kwunhang (Edwin) Wong,
Jiaming Lei,
Song Jin,
Shucheng Du,
Chuhan Zhang, \\
Songchen Ma, 
Weihao Zhang,
Jun Xiao,
and Kwang-Ting (Tim) Cheng$^{*}$%
\thanks{The authors are with The Hong Kong University of Science and Technology (HKUST),
ACCESS -- AI Chip Center for Emerging Smart Systems, and Zhejiang University.}%
\thanks{$^{*}$Corresponding author: Kwang-Ting (Tim) Cheng.}%
}

\begin{document}

\maketitle
\thispagestyle{empty}
\pagestyle{empty}

\begin{abstract}
World-action models (WAMs) have emerged as a promising paradigm for robot manipulation by jointly modeling future visual dynamics and robot actions. However, existing WAMs are trained predominantly on successful trajectories, making them prone to failure when real-world execution diverges from the learned dynamics. This issue is amplified in autoregressive WAMs, where execution errors become part of the causal history and continue to influence subsequent predictions. To this end, we introduce \method{}, a training-free framework that reformulates failure recovery as \emph{test-time scaling over causal histories}. This formulation decomposes recovery into three coupled decisions: \emph{when} to revise the causal history, \emph{where} to recover a reliable history prefix, and \emph{which} history configuration best supports subsequent execution. Specifically, \method{} realizes these decisions through three stages: 1) \textbf{Progress-Aware Recovery Trigger} detects persistent non-progress and triggers recovery only when the current execution state permits intervention; 2) \textbf{History-Prefix Recovery} identifies the unreliable history suffix, retrieves a historical anchor matching the current physical state, and reconstructs the causal KV state from the retained prefix while conditioning on the latest real observation; and 3) \textbf{Hypothesis Verification} compares the future continuations induced by complete-history, recovered-prefix, and full-reset hypotheses, and commits the best-supported hypothesis. Experiments in both simulated and real-world manipulation settings demonstrate consistent improvements in task success, while ablations confirm the contribution of each recovery stage.
\end{abstract}

\section{Introduction}

Vision-language-action (VLA) models have emerged as a scalable paradigm for general-purpose robot control, mapping visual observations and language instructions directly to executable actions~\cite{zitkovich2023rt,kim2024openvla,rss2025_avisionlanguagea,intelligence2025pi_}. By combining large-scale vision-language pretraining with robot demonstration data, these models support instruction-conditioned manipulation across diverse tasks and environments~\cite{brohan2022rt,o2024open,mees2024octo}. However, direct action prediction does not explicitly model how the physical scene is expected to evolve under the predicted action. Recently, world-action models (WAMs) couple action generation with predictive modeling of future visual states~\cite{du2023learning,li2025unifiedvideoactionmodel,zhu2025uwm,li2026causal,ye2026world}. Their imagined trajectories explicitly model future scene evolution, providing richer temporal context for long-horizon manipulation.

\begin{figure}[t]
    \centering
    \includegraphics[width=0.9\linewidth]{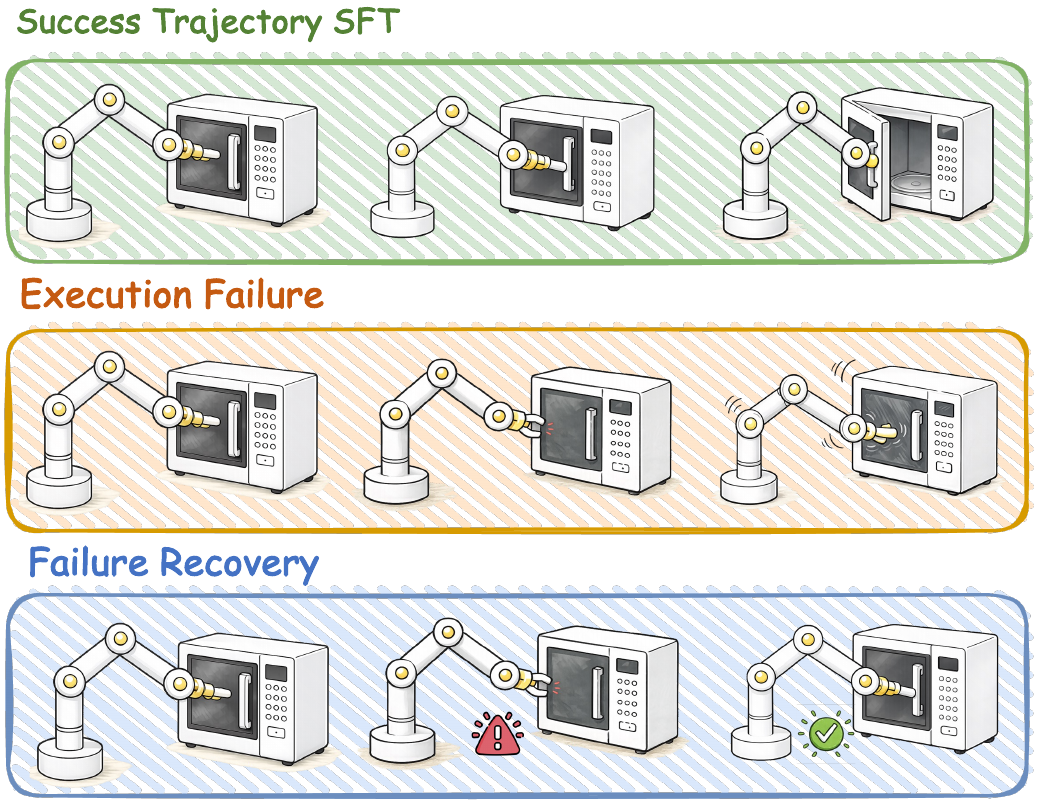}
    \vspace{-1.0em}
\caption{
\textbf{Failure recovery in autoregressive WAMs.} A model trained on successful trajectories learns a nominal grasp-and-open behavior (top). After a missed grasp, the retained causal history causes the policy to continue pulling and eventually oscillate near the closed door (middle). \method{} detects non-progress and revises the causal history for recovery (bottom).
}
    \label{fig:intro}
\end{figure}

Despite this predictive capability, existing WAMs are trained predominantly on successful robot demonstrations~\cite{li2026causal,ye2026world}, with failure states rarely represented during training. As illustrated in Fig.~\ref{fig:intro}, even a small localization or control error can cause the gripper to miss the microwave handle. Although the current observation reveals that the grasp has failed, the policy may continue the nominal pulling motion, eventually oscillating near the closed door rather than retrying the grasp. This problem is amplified in autoregressive WAMs such as LingBot-VA~\cite{li2026causal}, where visual and action tokens are retained in a persistent key--value (KV) cache across action chunks. Once execution diverges from the imagined trajectory, the resulting deviation remains encoded in the conditioning history and continues to influence subsequent predictions. Consequently, a transient execution error can persist through the causal history and repeatedly drive the model toward an invalid continuation.

Recent failure-recovery methods mainly differ in how corrective behavior is obtained. (i) \textit{Recovery-supervised methods} augment successful demonstrations with failed executions and recovery trajectories, enabling policies to learn corrective behaviors from off-nominal states~\cite{dai2025racer,zhao2026flare,yan2026redflow}. However, such recovery capability relies on additional recovery-specific supervision and policy optimization, and is inherently bounded by the failure patterns represented during training. (ii) \textit{Inference-time recovery methods} instead detect execution failures online and recover through symbolic replanning~\cite{namasivayam2024learning} or VLM-based corrective reasoning~\cite{chen2025robot}. These approaches typically rely on explicit state representations, predefined subgoals or skill libraries, and task-specific recovery interfaces, making them difficult to directly transfer to general autoregressive WAMs that jointly generate visual futures and action chunks.

In this work, we introduce \method{}, a training-free framework that reformulates failure recovery in autoregressive WAMs as \emph{test-time scaling over causal histories}. To the best of our knowledge, we are the first to explicitly treat the persistent KV-cached visual-action history of an autoregressive WAM as a recovery variable. Rather than scaling inference only over alternative futures conditioned on a fixed past~\cite{kwok2026scaling,sato2026sail}, \method{} treats the retained causal history itself as an additional test-time inference variable and evaluates the future continuations induced by alternative histories. This formulation gives rise to three coupled decisions: \emph{when} the retained history has become unreliable, \emph{where} the last reliable history prefix lies, and \emph{which} history hypothesis best supports subsequent execution.

To determine \emph{when} to revise the causal history, \method{} employs a \textbf{Progress-Aware Recovery Trigger} that monitors task progress using a frozen multi-view progress model~\cite{tan2025robo}. Recovery is triggered only after persistent non-progress is detected and the current execution state permits intervention. Otherwise, the base WAM follows its standard inference procedure, limiting additional test-time computation to cases where recovery is needed.

Once recovery is triggered, \textbf{History-Prefix Recovery} determines \emph{where} to revise the causal history. It identifies the unreliable history suffix, retrieves a historical anchor compatible with the current physical state, and reconstructs the causal KV state from the retained prefix using the WAM's native cache-construction procedure~\cite{li2026causal}. The reconstructed prefix is then paired with the latest real observation, preserving useful temporal context while grounding prediction in the current physical state.

Finally, \textbf{Hypothesis Verification} determines \emph{which} history hypothesis best supports subsequent execution. From a shared model state, \method{} evaluates complete-history continuation, prefix recovery, and full reset according to predicted task progress, visual continuity, and motion validity, and  commits the best-supported hypothesis to the persistent inference state. In this way, \method{} extends test-time scaling from alternative future actions or trajectories~\cite{kwok2026scaling,sato2026sail} to the causal histories that condition them, while keeping the underlying WAM fixed and requiring no recovery demonstrations or online policy optimization.

We evaluate \method{} on the RoboTwin~2.0 simulation benchmark~\cite{chen2025robotwin} and real-world robotic manipulation tasks. Extensive experiments demonstrate consistent improvements in task success across both settings, while detailed ablations validate the effectiveness of each recovery stage.

Our contributions are threefold:
\begin{itemize}
    \item We are the first to formulate failure recovery in autoregressive WAMs as \emph{test-time scaling over causal histories}, treating the persistent visual-action history encoded in the KV cache as an explicit recovery variable.
    \item We propose \method{}, a training-free recovery framework that determines \emph{when} to revise the causal history, \emph{where} to recover a reliable history prefix, and \emph{which} history hypothesis best supports subsequent execution.
    \item Extensive experiments on simulated and real-world manipulation tasks demonstrate the effectiveness of \method{}, with detailed ablations and visualizations validating the contributions of each stage.
\end{itemize}

\begin{figure*}[!t]
    \centering
    \includegraphics[width=1\linewidth]{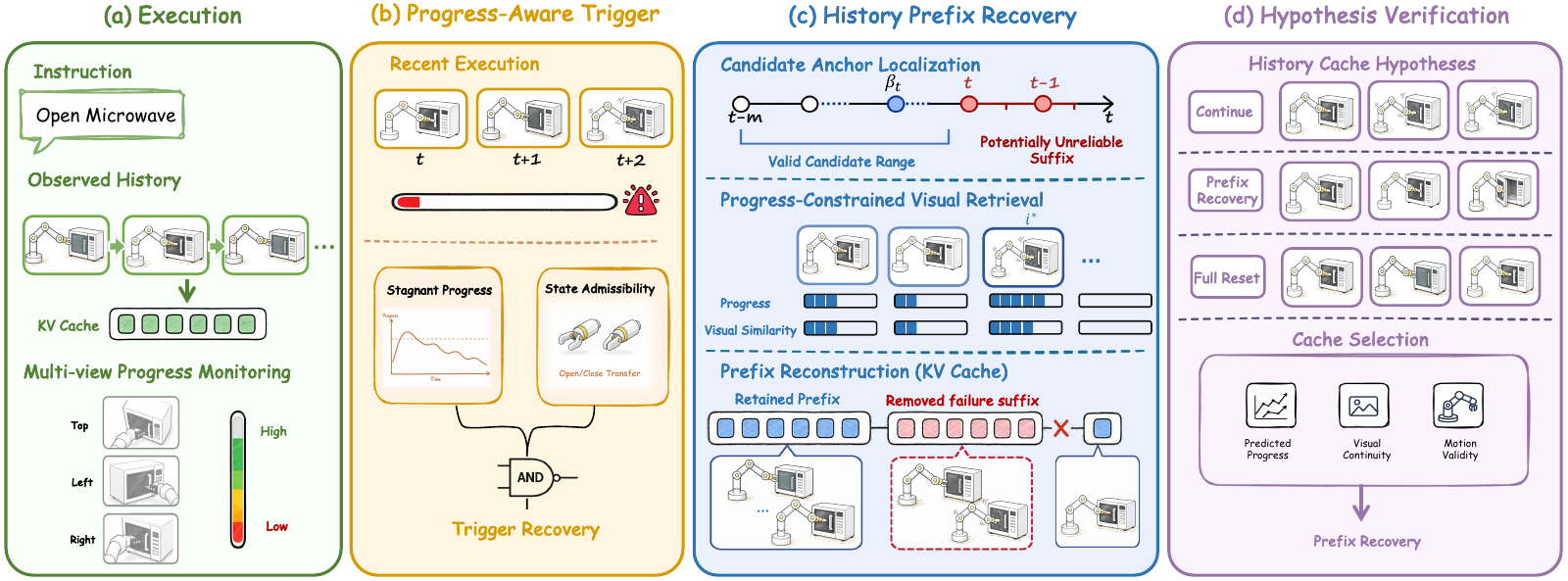}
    \vspace{-1.5em}
    \caption{
\textbf{Overview of the FaRe framework.}
\textbf{(a) Execution:} the autoregressive WAM performs closed-loop control while maintaining the causal history and monitoring realized progress.
\textbf{(b) Progress-Aware Recovery Trigger:} recovery is triggered upon persistent non-progress when the current execution state permits intervention.
\textbf{(c) History-Prefix Recovery:} a compatible historical anchor is retrieved to reconstruct the causal KV state and re-ground prediction on the latest real observation.
\textbf{(d) Hypothesis Verification:} continuation, prefix-recovery, and full-reset hypotheses are evaluated by predicted progress, visual continuity, and motion validity, with only the best-supported hypothesis committed.
}
    \label{fig:overview}
\end{figure*}

\section{Related Work}
\subsection{World-Action Models (WAMs)}
WAMs incorporate predictive visual dynamics into robot control and can be broadly grouped into two paradigms. The first follows an \emph{imagine-then-act} formulation, where future visual states or predictive representations are generated first and actions are subsequently inferred through inverse dynamics or a policy conditioned on the predicted future~\cite{du2023learning,tian2025predictive,hu2024video,li2026causal}. The second performs \emph{joint world-action modeling}, where future visual dynamics and robot actions are generated within a shared model or coupled generative process~\cite{kim2026cosmos,ye2026world,bi2026motus}. Recent work further explores how this coupling can be made more efficient, such as removing explicit future synthesis at inference time~\cite{yuan2026fast} or decoupling world planning from high-frequency action execution~\cite{cai2026aha}. Our work is complementary to these directions: rather than modifying how future worlds and actions are generated, we study failure recovery when the causal history retained by an autoregressive WAM becomes unreliable during execution.

\subsection{Test-Time Scaling (TTS)}
TTS improves a fixed policy by allocating additional inference computation to candidate generation, search, refinement, or verification. Recent robotic approaches mainly scale inference over alternative actions or trajectories. RoboMonkey samples multiple action proposals and selects among them with a learned verifier~\cite{kwok2025robomonkey}, while CoVer jointly scales instruction and action candidates through contrastive verification~\cite{kwok2026scaling}. TACO similarly samples multiple action chunks and ranks them using an inference-time verifier~\cite{yang2025steering}. At the trajectory level, SAIL performs search and iterative refinement over complete robot trajectories~\cite{sato2026sail}. Across these approaches, additional inference computation is primarily devoted to exploring or evaluating alternative future behaviors while the preceding interaction history remains fixed. Our work instead treats the causal history itself as an additional dimension of test-time scaling and evaluates the future continuations induced by alternative histories.

\subsection{Robot Failure Recovery}
Robot failure recovery can be broadly divided into learned recovery policies and inference-time corrective reasoning. The first line acquires corrective behavior through additional recovery experience or optimization.  RecoveryChaining learns local recovery policies with hierarchical reinforcement learning~\cite{vats2025recoverychaining}, RACER augments expert demonstrations with failure-recovery trajectories and language supervision~\cite{dai2025racer}, and FLARE introduces learned retry and reset mechanisms for execution deviations and state-breaking failures~\cite{zhao2026flare}. A second line performs failure diagnosis and correction during execution. Neuro-symbolic approaches localize erroneous plan states and search for valid subgoals~\cite{namasivayam2024learning}, while REFLECT~\cite{liu2023reflect} and VLM-based recovery methods~\cite{chen2025robot} use language or visual-language reasoning to diagnose failures and generate corrective plans or actions. These approaches primarily recover by revising future plans or actions. Nevertheless, autoregressive WAMs pose an additional challenge because executed visual-action history continues to condition future predictions. Our work instead revises and verifies causal histories at test time while keeping the base WAM fixed.

\section{Methodology}
\label{sec:method}

\noindent\textbf{Problem Formulation.}
We consider a frozen chunked autoregressive world-action model (WAM) $\pi_\theta$~\cite{li2026causal}. At policy step $t$, the model takes as input a language instruction $\ell$, the current multi-view observation $o_t=\{o_t^v\}_{v\in\mathcal{V}}$, and a persistent KV cache $K_t$ constructed from the executed visual-action history
$\mathcal{H}_{<t}=\left(o_0,(o_{i+1},\mathbf{a}_i)_{i=0}^{t-1}\right)$.
\footnote{Following the interleaved visual-action sequence~\cite{li2026causal}, $\mathcal{H}_{<t}$ represents the causal history of executed observations and action chunks, where each $\mathbf{a}_i$ is associated with the subsequent real observation $o_{i+1}$. The real observation is asynchronously updated into the history for the next policy step.}
Following inverse-dynamics modeling~\cite{li2026causal}, the WAM jointly models future visual evolution and action generation conditioned on the current observation and retained causal history. Specifically, it predicts an imagined visual trajectory $\hat{\mathbf{o}}_t=\{\hat{o}_{t,h}\}_{h=1}^{H}$ and an action chunk $\mathbf{a}_t=\{a_{t,h}\}_{h=1}^{H}$:
\begin{equation}
(\hat{\mathbf{o}}_t,\mathbf{a}_t)=\pi_\theta(o_t,\ell,K_t).
\label{eq:wam}
\end{equation}
After executing $\mathbf{a}_t$, the robot observes $o_{t+1}$, and a frozen progress reward model $R$~\cite{tan2025robo} evaluates the transition to provide an incremental progress estimate:
\begin{equation}
p_t=R(o_t,o_{t+1},\ell)\in[-1,1].
\label{eq:progress}
\end{equation}

Because the KV cache persists across action chunks, execution errors can introduce an unreliable suffix into the retained causal history, which continues to condition subsequent visual and action predictions. We therefore seek a history boundary $i<t$ such that the retained prefix $\mathcal{H}_{\le i}$ remains consistent with the current physical state and provides a reliable context for subsequent prediction.

Based on this formulation, we introduce \method{}, a training-free framework for test-time scaling over causal histories while keeping the underlying WAM fixed. As illustrated in Fig.~\ref{fig:overview}, \method{} determines \emph{when} to revise the causal history based on persistent non-progress and execution-state admissibility, \emph{where} to revise it by recovering a reliable history prefix, and \emph{which} history hypothesis best supports subsequent execution through hypothesis verification.

\subsection{Progress-Aware Recovery Trigger}
\label{sec:when}

To determine when to revise the causal history, \method{} monitors the realized task progress while accounting for the robot's current execution state. A single low-progress transition provides insufficient evidence for recovery, since temporary stagnation may arise from reward-model noise or reasonable contact-rich motions~\cite{tan2025robo,yan2026redflow}. In consequence, we trigger recovery only when persistent non-progress is detected and the current robot state permits intervention.

\paragraph{Accumulated task progress}
The progress reward model in Eq.~\eqref{eq:progress} provides a local estimate $p_t$ for each executed transition. However, directly accumulating these estimates would assign the same absolute change across different stages of task execution and may lead to an unbounded progress state~\cite{tan2025robo,yan2026redflow}. Thus, we integrate local progress using the following bounded relative update:
\begin{equation}
P_t=
\begin{cases}
P_{t-1}+(1-P_{t-1})p_t, & p_t\ge0,\\
P_{t-1}+P_{t-1}p_t, & p_t<0,
\end{cases}
\label{eq:hop_integration}
\end{equation}
where $P_0=0$ and $P_t\in[0,1]$ denotes the accumulated task progress. The update interprets $p_t$ as a relative change with respect to the current progress state. Positive estimates are scaled by the remaining progress $1-P_{t-1}$, moving the state toward completion, whereas negative estimates are scaled by the progress already attained $P_{t-1}$, moving it toward the lower progress boundary. Intuitively, this integration preserves the bounded range of $P_t$ while adapting the influence of each local estimate to the current stage of execution.

\paragraph{Persistent Stagnation}
Although $P_t$ captures overall task progress, isolated low-progress transitions do not necessarily indicate an execution failure. Thus, we identify persistent non-progress using two complementary temporal statistics. Let $s_t$ denote the number of consecutive steps ending at $t$ for which $p_i\le\epsilon_p$, measuring the duration of recent local stagnation. This criterion captures sustained short-term non-progress, but may not fully reflect slower failure modes in which the robot repeatedly oscillates without making net progress. To capture such behavior, we additionally track the running progress peak $P_t^\star=\max_{j\le t}P_j$ and let $k_t$ denote the number of consecutive steps for which accumulated progress fails to improve beyond the previous peak by more than $\epsilon_p$. The progress trigger is activated when $z_t=\mathbb{I}[s_t\ge N_s\lor k_t\ge N_k]$, where $N_s$ and $N_k$ control the persistence required for short-term and longer-term stagnation, respectively. The two criteria complement each other by capturing both consecutive local non-progress and prolonged failure to exceed previously achieved progress. Nevertheless, persistent stagnation alone does not imply that immediate intervention is physically appropriate. Therefore, we further assess whether the current robot state is suitable for recovery.

\paragraph{State Admissibility}
Intuitively, states around grasp establishment or gripper-state transitions, such as closing the gripper on an object or switching between open and closed states, often correspond to active manipulation phases in which interrupting execution may disrupt an otherwise valid interaction. Therefore, we assess state admissibility from the recent proprioceptive dynamics of the arm and gripper. Specifically, for each arm $b$, let $M_{t,b}$ and $G_{t,b}$ denote the arm displacement and gripper variation over the preceding action chunk, respectively. Intervention is deferred when substantial arm motion ($M_{t,b}>\tau_a$) coincides with significant gripper variation ($G_{t,b}>\tau_g$), indicating that the robot may still be undergoing an active grasp or contact transition. Accordingly, the admissibility gate $g_t\in\{0,1\}$ is set to one only when no arm simultaneously satisfies these conditions. Recovery is activated when both persistent stagnation and state admissibility are satisfied, \ie, $r_t=z_tg_t=1$. Otherwise, the base WAM continues its nominal inference path and the recovery condition is reconsidered at the next action-chunk boundary.

\subsection{History-Prefix Recovery}
\label{sec:where}

Once recovery is triggered, \method{} determines \emph{where} to revise the causal history. The goal is not simply to roll back by a fixed number of steps, since the duration of an execution failure may vary substantially across episodes. Instead, we seek a history prefix that remains consistent with the achieved task progress while being compatible with the current physical scene. To this end, \method{} first uses realized progress to localize a candidate recovery region and then retrieves the historical state that best matches the current physical configuration using multi-view visual evidence.

\paragraph{Candidate Anchor Localization}
Specifically, we first identify the most recent transition with meaningful positive progress, $\beta_t=\max\{i<t\mid p_i>\epsilon_p\}$.  Since $p_i$ measures the progress of the transition from $o_i$ to $o_{i+1}$, $\beta_t$ provides a progress-based boundary for locating the reliable history. We restrict candidate anchor indices to $i\le\beta_t$ and require them to be at least $r_{\min}$ action chunks away from the current step. The history after each candidate anchor is treated as potentially unreliable. This progress-based constraint adapts the recovery horizon to the observed execution, allowing \method{} to search farther back when non-progress persists across multiple action chunks.

\paragraph{Progress-Constrained Visual Retrieval}
Progress localization restricts the search to a task-consistent stage, but does not guarantee that the corresponding historical state matches the current physical configuration. Conversely, visual similarity alone may retrieve a state from an inappropriate stage of the task. We therefore perform multi-view visual retrieval within the progress-constrained candidate set. For a candidate transition $i$, the corresponding post-transition observation is $o_{i+1}$. Given a reference observation $\bar{o}_t$ constructed from recent real observations, we define its visual discrepancy as:
\begin{equation}
d_i = \max_{v\in\mathcal{V}} \operatorname{MSE}\left(\bar{o}_t^v,o_{i+1}^v\right).
\label{eq:visual_distance}
\end{equation}
Taking the maximum across views penalizes candidates inconsistent in any viewpoint. We discard candidates with $d_i>\tau_v$ and select the anchor $i^*$ with the smallest visual discrepancy, using recency only for tie-breaking. Thus, progress restricts retrieval to the appropriate task stage, while multi-view evidence identifies the best-matching historical state.

\begin{figure}
    \centering
    \includegraphics[width=1\linewidth]{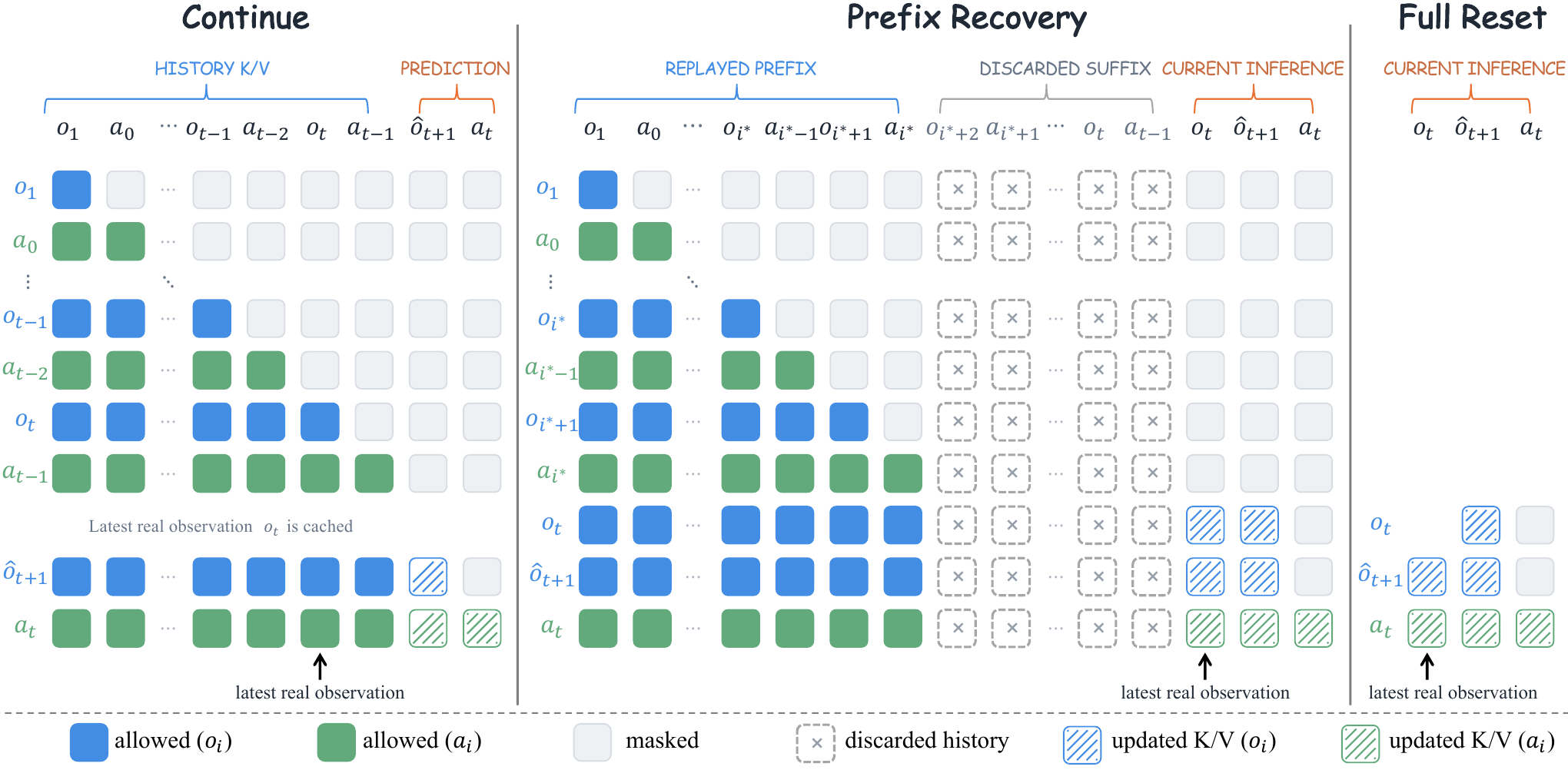}
    \vspace{-2.0em}
    \caption{\textbf{History-cache hypotheses in \method{}.}
\textsc{Continue} retains the complete persistent KV cache, \textsc{Prefix} reconstructs the cache from a reliable history prefix while removing the unreliable suffix, and \textsc{Reset} discards the persistent history entirely.
All hypotheses share the same latest real observation $o_t$ and frozen WAM~\cite{li2026causal}.}
\label{fig:cache_hypotheses}
    \label{fig:kvcache}
\end{figure}

\paragraph{Causal Prefix Reconstruction}
Once the anchor $i^*$ is selected, \method{} reconstructs the causal KV state by replaying the retained history prefix through the WAM's native cache-construction procedure~\cite{li2026causal}. Following the native interleaved ordering illustrated in Fig.~\ref{fig:cache_hypotheses}, the replayed prefix retains the history through $o_{i^*+1}$ and $\mathbf{a}_{i^*}$, while the subsequent suffix is discarded, yielding $K_t^{\mathrm{pre}}=\mathcal{C}(\mathcal{H}_{\le i^*})$. This reconstruction preserves the WAM's native positional structure and visual-action ordering while removing the potentially unreliable suffix. The recovered prefix is then conditioned on the latest real observation $o_t$:
\begin{equation}
(\hat{\mathbf{o}}_t^{\mathrm{pre}},\mathbf{a}_t^{\mathrm{pre}})=\pi_\theta\left(o_t,\ell,K_t^{\mathrm{pre}}\right).
\label{eq:prefix_rollout}
\end{equation}
Conditioning the recovered prefix on the latest real observation preserves task-relevant temporal context while grounding prediction in the current physical state. As shown in Fig.~\ref{fig:cache_hypotheses}, only the unreliable history suffix is removed, while $o_t$ remains the current input under the WAM's native causal attention structure.

\subsection{Hypothesis Verification}
\label{sec:whether}

The recovered prefix provides one possible revision of the causal history, but it may not always be preferable. When the execution deviation is mild, retaining the complete history can preserve useful context, whereas larger deviations may require discarding the persistent history entirely. We therefore determine \emph{which} history hypothesis best supports subsequent execution by comparing the future continuations induced by different causal histories.

\paragraph{History-Cache Hypotheses}
When recovery is triggered, \method{} considers three cache hypotheses corresponding to different degrees of causal-history revision, as illustrated in Fig.~\ref{fig:cache_hypotheses}. \textsc{Continue} retains the complete persistent history and serves as the no-revision hypothesis, allowing the original execution to proceed when history revision is unnecessary or the existing context remains reliable. \textsc{Prefix} uses the reconstructed cache $K_t^{\mathrm{pre}}$ from Sec.~\ref{sec:where}, retaining the reliable prefix while discarding the potentially unreliable suffix. \textsc{Reset} removes the persistent history entirely and predicts from the current observation alone. All three hypotheses use the same current observation $o_t$, instruction $\ell$, and frozen WAM, differing only in the retained history cache. For each hypothesis $q$, we generate $(\hat{\mathbf{o}}_t^q,\mathbf{a}_t^q)=\pi_\theta(o_t,\ell,K_t^q)$,
allowing the effect of different causal histories to be compared through their predicted continuations.

\paragraph{Verification Score}
A preferred recovery hypothesis should predict task advancement while remaining visually grounded in the current physical state and producing a meaningful action response. Accordingly, we evaluate each history-cache hypothesis using three complementary criteria: prospective task progress, visual continuity, and motion validity. For hypothesis $q$, we estimate its predicted progress as $\hat{p}_q=R(o_t,\hat{o}_{t,H}^q,\ell)$. However, only utilizing progress may favor a rollout whose imagined future is inconsistent with the current observation or whose predicted action is nearly static~\cite{ruan2026future}. Thus, we define $e_q=\max_{v\in\mathcal{V}}\operatorname{MSE}(o_t^v,\hat{o}_{t,1}^{q,v})$ as the visual discrepancy between the current observation and the first imagined state, and let $m_q$ denote the maximum arm displacement within the predicted action chunk. Each hypothesis is scored as:
\begin{equation}
J_q = \hat{p}_q - \lambda_c e_q - \lambda_m \frac{[\tau_m-m_q]_+}{\tau_m},
\label{eq:candidate_score}
\end{equation}
where $[x]_+=\max(0,x)$. The visual term penalizes continuations that are poorly grounded in the current scene. The motion term penalizes action chunks whose arm displacement falls below the minimum effective-motion threshold $\tau_m$, with the normalization by $\tau_m$ bounding this penalty between zero and one. We select the highest-scoring hypothesis as $q^*=\arg\max_q J_q$. To prevent candidate evaluation from altering the persistent inference state, all hypotheses are evaluated from a shared model snapshot and their temporary cache updates are discarded after scoring. The selected hypothesis is then regenerated from the same snapshot, yielding the executed action chunk $\mathbf{a}_t=\mathbf{a}_t^{q^*}$ and its corresponding KV state, which is committed to persistent memory.

\section{Experiments}
We evaluate \method{} in both RoboTwin~2.0 simulation and real-world manipulation to study the effectiveness of test-time causal-history revision for failure recovery in autoregressive WAMs. Our experiments address three key questions: 1) Can \method{} improve task success and recover from diverse execution failures without recovery training or parameter updates? 2) How do the recovery trigger, history-prefix recovery, and hypothesis verification contribute to the overall performance? 3) Can the same framework generalize from simulation to real-world robot manipulation?

\subsection{Experimental Setup}
We evaluate our \method{} in both simulated and real-world manipulation settings. For simulation, we use RoboTwin~2.0~\cite{chen2025robotwin} as the primary benchmark and report task success rate (SR) under the Easy and Hard evaluation settings. For real-world experiments, we deploy the policy on a dual-arm SO-101 robot platform~\cite{cadene2026lerobot} and evaluate three manipulation tasks: \emph{Fold Towel}, \emph{Stack C-shaped Blocks}, and \emph{Organize Vegetables}. Each task is independently executed for 20 trials, and we report the corresponding empirical success rate as the primary evaluation metric. Unless otherwise specified, the same model checkpoints and recovery hyperparameters are used across all evaluation tasks.

\subsection{Implementation Details}

We leverage the released checkpoint post-trained on RoboTwin~2.0~\cite{li2026causal} and Robo-Dopamine GRM-2.0-8B-Preview~\cite{tan2025robo} as the frozen progress model. Both progress estimation and history retrieval use the top, left, and right RoboTwin camera views. For the recovery trigger in Sec.~\ref{sec:when}, we set $\epsilon_p=0.03$, $N_s=2$, and $N_k=3$, with arm-motion and gripper-variation thresholds $\tau_a=10^{-4}$ and $\tau_g=0.002$, respectively. For history-prefix recovery in Sec.~\ref{sec:where}, $\bar{o}_t$ is constructed by averaging the two most recent real observations, with $r_{\min}=2$ and $\tau_v=0.06$. As for hypothesis verification (Sec.~\ref{sec:whether}), each candidate spans one action chunk, with $\lambda_c=2.0$, $\lambda_m=0.10$, and $\tau_m=0.01$ in Eq.~\eqref{eq:candidate_score}. All hypothesis candidate rollouts are evaluated from a shared model snapshot, and the selected hypothesis is regenerated before executing its action chunk and committing the corresponding KV state. All experiments are conducted on NVIDIA H20 GPUs.

\subsection{Main Results}

\subsubsection{Results on RoboTwin~2.0}
We report the success rates of state-of-the-art methods under the \textit{Easy} and \textit{Hard} settings in Table~\ref{tab:robotwin_main}. As seen, \method{} improves the reimplemented LingBot-VA baseline from 91.5\% to 93.2\% on Easy and from 90.2\% to 92.4\% on Hard, yielding absolute gains of 1.7\% and 2.2\%, respectively. Despite the strong baseline performance, these improvements correspond to relative reductions of 20.0\% and 22.4\% in failure rate. Importantly, the gains are achieved without additional recovery demonstrations or parameter updates to the underlying WAM. The consistent gains across both settings show that test-time causal-history revision can further improve a strong frozen autoregressive WAM without additional training.

\subsubsection{Results on Real-World Manipulation}
Table~\ref{tab:real_robot} shows the results on the dual-arm SO-101~\cite{cadene2026lerobot} platform. \method{} improves the average success rate from 71.7\% to 83.3\%, with consistent gains across all three tasks. Specifically, the success rate increases from 90.0\% to 100.0\% on \emph{Fold Towel}, from 65.0\% to 80.0\% on \emph{Stack C-shaped Blocks}, and from 60.0\% to 70.0\% on \emph{Organize Vegetables}. The largest improvement is observed on \emph{Stack C-shaped Blocks}, with an absolute gain of 15.0\%. Overall, \method{} succeeds in 50 of 60 trials, compared with 43 for the base policy. These results show that the proposed recovery mechanism remains effective under real-world execution and is not limited to simulated environments.

\begin{table}[t]
\centering
\caption{\textbf{Quantitative Results on RoboTwin~2.0.}
``$\ast$'' denotes the reimplementation, while the other baseline results are taken from prior work.
TTS indicates whether test-time scaling is applied.}
\label{tab:robotwin_main}
\renewcommand{\arraystretch}{1.05}
\setlength{\tabcolsep}{10pt}
\begin{tabular}{lccc}
\toprule
\multirow{2}{*}{\textbf{Method}}
& \multirow{2}{*}{\textbf{TTS}}
& \multicolumn{2}{c}{\textbf{Average SR (\%)}} \\
\cmidrule(lr){3-4}
& & \textbf{Easy} & \textbf{Hard} \\
\midrule
X-VLA~\cite{zheng2026x}
& $\times$ & 72.8 & 72.8 \\

$\pi_0$~\cite{rss2025_avisionlanguagea}
& $\times$ & 65.9 & 58.4 \\

$\pi_{0.5}$~\cite{intelligence2025pi_}
& $\times$ & 82.7 & 76.8 \\

ABot-M0~\cite{yang2026abotm0vlafoundationmodel}
& $\times$ & 81.2 & 80.4 \\

Motus~\cite{bi2026motus}
& $\times$ & 88.7 & 87.0 \\

\hdashline
$\ast$LingBot-VA~\cite{li2026causal}
& $\times$ & 91.5 & 90.2 \\

\textbf{+ \method{} (ours)}
& $\checkmark$ & \textbf{93.2} & \textbf{92.4} \\
\bottomrule
\end{tabular}
\end{table}

\begin{table}[t]
\centering
\caption{\textbf{Ablation of the main components of \method{} on RoboTwin~2.0.}}
\label{tab:ablation_components}
\renewcommand{\arraystretch}{1.0}
\setlength{\tabcolsep}{7pt}
\begin{tabular}{lccc}
\toprule
\textbf{Variant}
& \textbf{Average SR (\%)} \\
\midrule
LingBot-VA
& 91.5 \\

\textbf{\method{}}
& \textbf{93.2}   \\
\hdashline
\textit{w/o} Recovery Trigger
& 91.0  \\
\textit{w/o} Prefix Recovery
& 92.5 \\
\textit{w/o} Hypothesis Verification
& 91.2 \\
\bottomrule
\end{tabular}
\end{table}

\begin{table}[t]
\centering
\caption{\textbf{Real-world manipulation results on the dual-arm SO-101 platform.} Each task is evaluated over 20 independent trials, and the success rate is reported.}
\label{tab:real_robot}
\renewcommand{\arraystretch}{1.0}
\setlength{\tabcolsep}{4pt}
\begin{tabular}{lcccc}
\toprule
\textbf{Method} & \textbf{Towel} & \textbf{Blocks} & \textbf{Vegetables} & \textbf{Avg.} \\
\midrule
LingBot-VA~\cite{li2026causal} & 90.0 & 65.0 & 60.0 & 71.7 \\
\textbf{+ \method{}} & \textbf{100.0} & \textbf{80.0} & \textbf{70.0} & \textbf{83.3} \\
\bottomrule
\end{tabular}
\end{table}

\begin{table}[t]
\centering
\caption{\textbf{Ablation of history-cache hypotheses on RoboTwin~2.0.}}
\label{tab:ablation_hypotheses}
\renewcommand{\arraystretch}{1.0}
\setlength{\tabcolsep}{7pt}
\begin{tabular}{cccc}
\toprule
\textsc{Continue}
& \textsc{Prefix}
& \textsc{Reset}
& {\textbf{Average SR (\%)}} \\ 
\midrule
$\checkmark$ & $\times$     & $\times$     & 91.5  \\
$\checkmark$ & $\checkmark$ & $\times$     & 92.6  \\
$\checkmark$ & $\times$     & $\checkmark$ & 92.5  \\
$\checkmark$ & $\checkmark$ & $\checkmark$
& \textbf{93.2} \\
\bottomrule
\end{tabular}
\end{table}

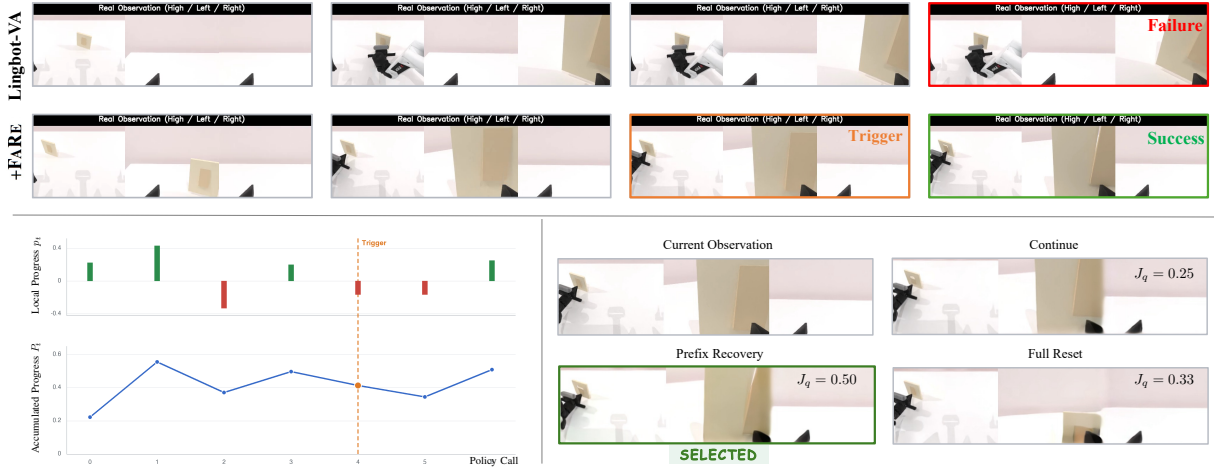
\begin{figure*}
    \centering
    \includegraphics[width=0.9\linewidth]{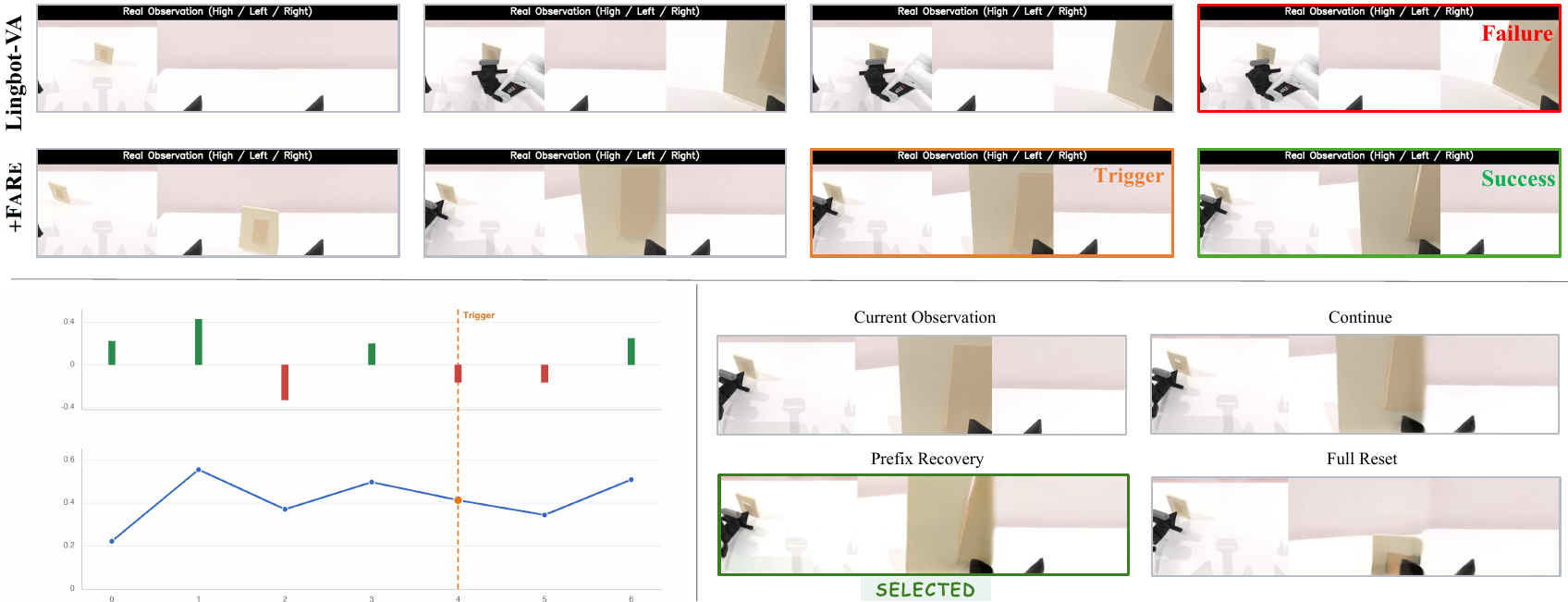}
    \put(-157, 30){\scalebox{0.55}{$J_q=0.50$}}
    \put(-30, 70){\scalebox{0.55}{$J_q=0.25$}}
    \put(-30, 30){\scalebox{0.55}{$J_q=0.33$}}
    \put(-445,70){\rotatebox[origin=c]{90}{\scalebox{0.4}{Local Progress $p_t$}}}
    \put(-445,24){\rotatebox[origin=c]{90}{\scalebox{0.4}{Accumulated Progress $P_t$}}}
    \put(-280, 0){\scalebox{0.4}{Policy Call}}
    \caption{
\textbf{Qualitative comparison and recovery-hypothesis verification.}
\textbf{Top:} Executed trajectories of LingBot-VA and \method{}, ending in failure and success, respectively.
\textbf{Bottom left:} Local progress $p_t$ and accumulated progress $P_t$, with the dashed line marking recovery activation.
\textbf{Bottom right:} The current observation and predicted continuations under \textsc{Continue}, \textsc{Prefix}, and \textsc{Reset}.
The prefix-recovery hypothesis receives the highest verification score and is selected for execution.
All hypotheses share the latest real observation; only the retained conditioning history is revised.
}
    \label{fig:quali}
\end{figure*}

\makeatletter
\begin{figure*}[t]
    \centering

    \begin{minipage}[t]{0.34\textwidth}
        \vspace{0pt}
        \centering
        \def\@captype{table}

        \caption{\textbf{Ablation study of verification criteria on RoboTwin~2.0.}}
        \label{tab:ablation_verification}

        {\footnotesize
        \renewcommand{\arraystretch}{1.0}
        \setlength{\tabcolsep}{3.5pt}
        \begin{tabular}{cccc}
            \toprule
            \textbf{Progress}
            & \textbf{Visual}
            & \textbf{Motion}
            & \textbf{Avg. SR (\%)} \\
            \midrule
            $\checkmark$ & $\times$     & $\times$     & 92.4 \\
            $\checkmark$ & $\checkmark$ & $\times$     & 92.9 \\
            $\checkmark$ & $\times$     & $\checkmark$ & 92.6 \\
            $\checkmark$ & $\checkmark$ & $\checkmark$ & \textbf{93.2} \\
            \bottomrule
        \end{tabular}
        }

        \vspace{1.0em}

        \caption{\textbf{Comparison of action-chunk inference latency.}}
        \label{tab:latency}

        {\footnotesize
        \renewcommand{\arraystretch}{1.0}
        \setlength{\tabcolsep}{4pt}
        \begin{tabular}{lc}
            \toprule
            \textbf{Method}
            & \textbf{Latency (ms/chunk)} $\downarrow$ \\
            \midrule
            LingBot-VA~\cite{li2026causal}
            & 6200.7 \\
            +\method{} (ours)
            & 6872.4 \\
            \bottomrule
        \end{tabular}
        }
    \end{minipage}
    \hfill
    \begin{minipage}[t]{0.63\textwidth}
        \vspace{0pt}
        \centering
        \def\@captype{figure}
        \includegraphics[width=\linewidth]{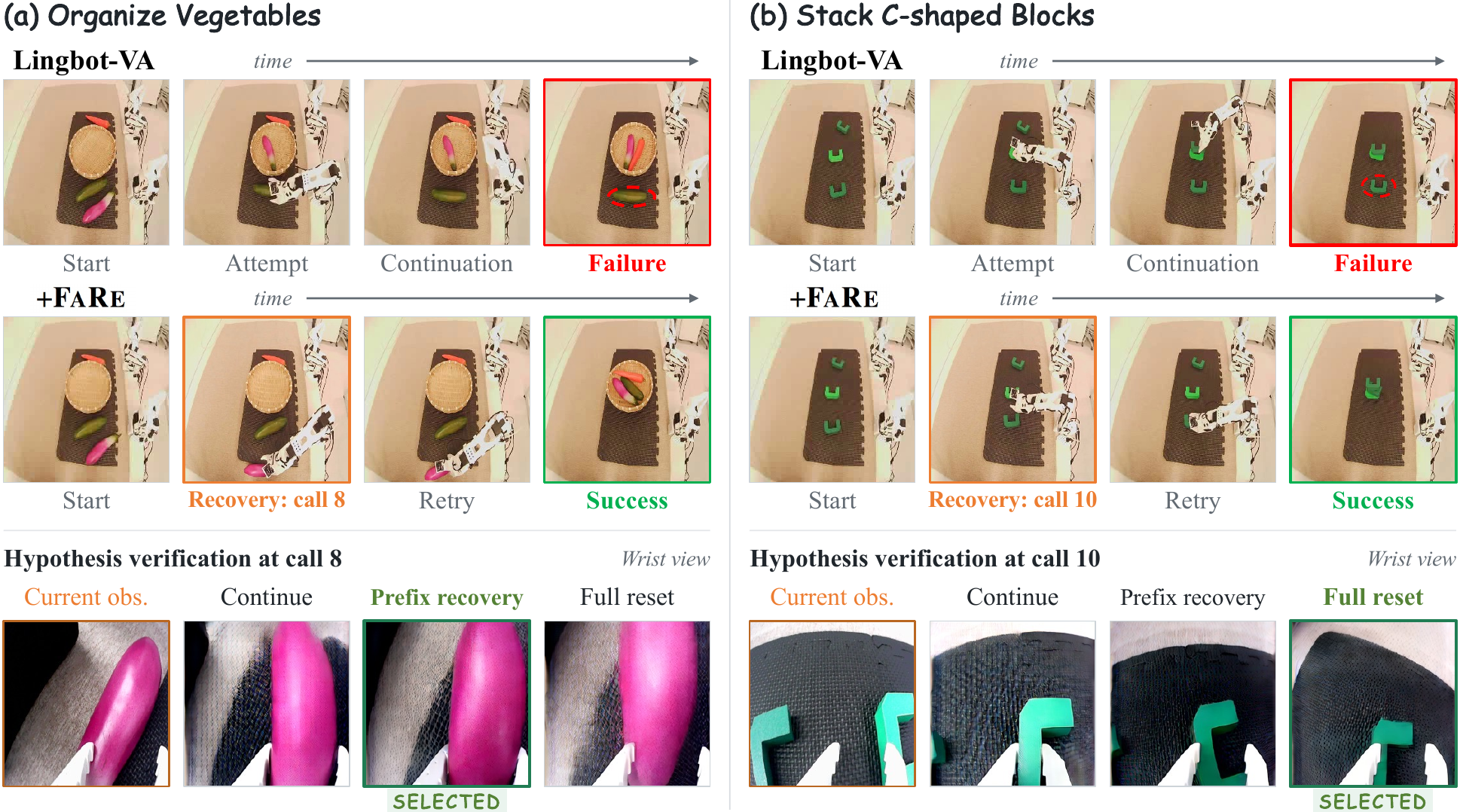}
        \vspace{-1.5em}
        \caption{\textbf{Real-world failure recovery and hypothesis verification.}}
        \label{fig:real_world_qualitative}
    \end{minipage}

\end{figure*}
\makeatother

\subsection{Ablation Studies}

\subsubsection{Effectiveness of Each Component}
We ablate each of the three recovery stages while keeping the remaining components unchanged, as summarized in Table~\ref{tab:ablation_components}. 
\textbf{(i) w/o Recovery Trigger} generates and verifies recovery hypotheses at every decision step. The success rate drops from 93.2\% to 91.0\%, even below the LingBot-VA baseline, indicating that more frequent recovery search does not necessarily improve execution and highlighting the importance of selective intervention. \textbf{(ii) w/o Prefix Recovery} removes the reconstructed-prefix hypothesis and retains only \textsc{Continue} and \textsc{Reset}, reducing the success rate to 92.5\%. This result shows that partial history revision provides a useful alternative to either retaining or discarding the entire causal history. \textbf{(iii) w/o Hypothesis Verification} directly commits the recovered prefix when a valid anchor is found and otherwise performs a full reset. The success rate decreases to 91.2\%, demonstrating that a retrieved prefix should be evaluated through its induced continuation before being committed. Overall, these results confirm the complementary roles of recovery triggering, prefix recovery, and hypothesis verification.

\subsubsection{Choice of History-Cache Hypotheses}
Table~\ref{tab:ablation_hypotheses} studies the effect of the history-cache hypotheses (\cf~\ref{sec:whether}) used during verification while keeping the trigger, retrieval procedure, and scoring function fixed. \textsc{Continue} alone achieves 91.5\% success. Adding \textsc{Prefix} improves the success rate to 92.6\%, while adding \textsc{Reset} yields 92.5\%. Using all three hypotheses achieves the best performance of 93.2\%. These results indicate that prefix recovery and full reset provide complementary forms of causal-history revision: the former preserves useful historical context, whereas the latter allows recovery without persistent context when the retained history is no longer reliable.

\subsubsection{Effectiveness of the Verification Criteria}
Table~\ref{tab:ablation_verification} evaluates the individual terms in Eq.~\eqref{eq:candidate_score} while keeping the hypothesis set fixed. Using predicted progress alone achieves 92.4\% success. Adding visual continuity increases the success rate to 92.9\%, while adding motion validity yields 92.6\%. Combining all three criteria achieves the best performance of 93.2\%. The visual term favors continuations that remain consistent with the current observation, while the motion term suppresses nearly static action chunks. These results show that visual grounding and motion validity provide complementary signals beyond predicted task progress for selecting recovery hypotheses.

\subsubsection{Inference Latency}
Table~\ref{tab:latency} reports the average inference latency over 1K action chunks. \method{} increases the per-chunk latency from 6200.7 ms to 6872.4 ms. The three recovery hypotheses are evaluated in parallel, while progress-model inference is accelerated using vLLM~\cite{kwon2023efficient}. Since hypothesis verification is invoked only at recovery steps, with at most five such interventions per episode, the additional computation remains limited, keeping the overall inference overhead modest.

\subsection{Qualitative Analysis}
\subsubsection{Simulated Scenario}
Fig.~\ref{fig:quali} shows a recovery episode where the base policy fails, while \method{} recovers and completes the task. Before intervention, intermittent positive progress does not raise the accumulated progress beyond its previous peak, indicating persistent stagnation. At the recovery trigger, the three history hypotheses are evaluated from the same current observation. \textsc{Prefix} obtains the highest verification score ($J_q=0.50$), compared with \textsc{Continue} ($0.25$) and \textsc{Reset} ($0.33$), and is selected for execution. This example shows that retaining a compatible prefix can preserve useful context while removing the unreliable failure suffix. Recovery only revises the conditioning cache, with prediction remaining grounded in the latest real observation.

\subsubsection{Real-World Scenario}
Fig.~\ref{fig:real_world_qualitative} presents two real-world recovery episodes. As seen, Lingbot-VA~\cite{li2026causal} leaves a vegetable outside the basket in (a) and produces an incomplete block stack in (b), while \method{} successfully completes both tasks after recovery. The selected hypotheses illustrate complementary forms of causal-history revision: \textsc{Prefix} preserves compatible context while removing the unreliable suffix in (a), whereas \textsc{Reset} discards the persistent history and replans from the current observation in (b). These episode-dependent choices show that verification adaptively determines how much historical context to retain. These examples highlight the complementary roles of prefix recovery and full reset, while showing that verification of their induced continuations enables effective test-time recovery with a fixed WAM.

\section{Conclusion}

We presented \method{}, a training-free framework for failure recovery in autoregressive WAMs, formulated as \emph{test-time scaling over causal histories}. \method{} determines \emph{when} to revise the causal history through progress-aware triggering, \emph{where} to recover a reliable history prefix through progress-constrained retrieval and causal cache reconstruction, and \emph{which} history hypothesis to execute through verification over continuation, prefix-recovery, and full-reset alternatives. Experiments on RoboTwin~2.0 and real-world dual-arm SO-101 manipulation demonstrate the effectiveness of causal-history revision for recovering from execution failures, while ablations validate the contributions of the recovery trigger, history-prefix recovery, and hypothesis verification. More broadly, our results suggest that the conditioning history itself provides an additional dimension for test-time scaling in autoregressive world models, beyond scaling only over alternative future actions or trajectories.

\bibliographystyle{IEEEtran}
\bibliography{IEEEfull}

\end{document}